\documentclass{article}

\usepackage{amsmath}

    \usepackage[preprint]{neurips_2024}

\usepackage{multirow}
\usepackage[table,xcdraw]{xcolor}
\usepackage[utf8]{inputenc} 
\usepackage[T1]{fontenc}    
\usepackage{hyperref}       
\usepackage{url}            
\usepackage{booktabs}       
\usepackage{amsfonts}       
\usepackage{nicefrac}       
\usepackage{microtype}      
\usepackage{xcolor}         
\usepackage{graphicx}
\usepackage{amsmath}
\usepackage{comment}
\usepackage{algorithm}
\usepackage{algorithmicx}
\usepackage{diagbox}
\usepackage{multirow}
\usepackage{algpseudocode}
\usepackage{ifthen}
\usepackage{hyperref}
\usepackage{url}
\usepackage{booktabs}       
\usepackage{amsfonts}       
\usepackage{nicefrac}       
\usepackage{microtype}      
\usepackage{xcolor}         
\usepackage{amsmath}
\usepackage{bbm}
\usepackage{ulem}
\usepackage{graphicx}       
\usepackage{subcaption}
\usepackage{wrapfig}
\usepackage{makecell}
\usepackage{float}
\usepackage{appendix}
\usepackage{algorithm}
\usepackage{algorithmicx}
\usepackage{diagbox}

\usepackage{xcolor}         
\usepackage{graphicx} 
\usepackage[table,xcdraw]{xcolor}
\usepackage{authblk}

\title{MambaLLIE: Implicit Retinex-Aware Low Light Enhancement with Global-then-Local State Space}

\author[1]{Jiangwei Weng}
\author[1]{Zhiqiang Yan}
\author[2]{Ying Tai}
\author[1]{Jianjun Qian}
\author[1]{Jian Yang}
\author[1]{Jun Li}
\affil[1]{Nanjing University of Science and Technology, Nanjing, China}
\affil[2]{Nanjing University, Nanjing, China}

\begin{document}

\maketitle


\begin{abstract}
   Recent advances in low light image enhancement have been dominated by Retinex-based learning framework, leveraging convolutional neural networks (CNNs) and Transformers. However, the vanilla Retinex theory primarily addresses global illumination degradation and neglects local issues such as noise and blur in dark conditions. Moreover, CNNs and Transformers struggle to capture global degradation due to their limited receptive fields. While state space models (SSMs) have shown promise in the long-sequence modeling, they face challenges in combining local invariants and global context in visual data.
   In this paper, we introduce MambaLLIE, an implicit Retinex-aware low light enhancer featuring a global-then-local state space design. We first propose a Local-Enhanced State Space Module (LESSM) that incorporates an augmented local bias within a 2D selective scan mechanism, enhancing the original SSMs by preserving local 2D dependency. Additionally, an Implicit Retinex-aware Selective Kernel module (IRSK) dynamically selects features using spatially-varying operations, adapting to varying inputs through an adaptive kernel selection process. Our Global-then-Local State Space Block (GLSSB) integrates LESSM and IRSK with LayerNorm as its core. This design enables MambaLLIE to achieve comprehensive global long-range modeling and flexible local feature aggregation. Extensive experiments demonstrate that MambaLLIE significantly outperforms state-of-the-art CNN and Transformer-based methods. \href{https://mamballie.github.io/anon/}{Project Page}.

\end{abstract}

\section{Introduction}
\label{Introduction}

low light image enhancement is a challenging task in computer vision due to insufficient lighting and sensor degradation. Consequently, images often suffer from poor global visibility and local issues such as color distortion and noise. These degraded images can adversely affect human perception and high-level vision tasks, such as object detection.

Traditional techniques, including histogram equalization \cite{HistogramEqualization} and gamma correction \cite{gammacorrection}, enhance images through global mapping operations. However, these global operations often struggle to address local degradation effectively.
In recent years, many methods based on CNNs and Transformers have gradually come to dominate this field \cite{RetinexNet,KinD,ZeroDCE,SCI,SNR-Net,RetinexFormer}.
CNN-based methods \cite{RetinexNet,KinD,ZeroDCE,SCI,URetinex} have achieved significant advancements by effectively aggregating local information, thus substantially improving performance in low light enhancement. Nevertheless, the limited receptive field and weight-sharing strategy of CNNs result in a local reductive bias, making the models less adaptive to varying inputs.
On the other hand, Transformer-based methods \cite{SNR-Net,RetinexFormer,Restormer} achieve a larger and adaptive receptive field by emphasizing long-term dependencies through the self-attention mechanism. However, the vanilla attention mechanism scales quadratically with input size, resulting in significant computational overhead.

\begin{figure}[t]
  \centering
  \includegraphics[width=1\textwidth]{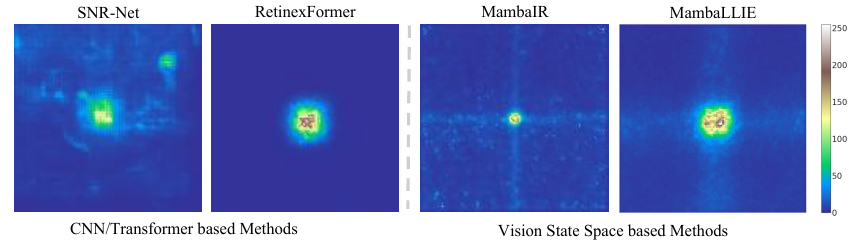}
  \caption{The Effective Receptive Field (ERF) visualization \cite{ERP} for SNR-Net \cite{SNR-Net}, RetinexFormer \cite{RetinexFormer}, MambaIR \cite{MambaIR} and our MambaLLIE. A broader distribution of bright areas signifies a larger ERF. The receptive field of SNR-Net is large but messy, due to the SNR-aware mechanism, RetinexFormer achieves a larger receptive field of the central point, and MambaIR has the the global receptive field, but presents the limited local perception. Only our proposed MambaLLIE achieves a global perception ability outwards from central point and preserves the large local receptive field.}
  \label{fig:erf}
  \vspace{-0.2in}
\end{figure}

Recently, Mamba\cite{Mamba,Vmamba,Videomamba} have garnered significant attention in the field of computer vision. These internal state space models (SSMs) demonstrate great potential for global information modeling with linear complexity. However, a straightforward implementation of vision state space models for low light image enhancement is inadequate. This is because SSMs are primarily designed for long-range modeling and lack the flexibility to capture local information effectively \cite{AugmentedTransformer}.
For instance, as illustrated in Fig. \ref{fig:erf}, the receptive field of MambaIR \cite{MambaIR}, a simple yet effective vision state space model, achieves longer-range dependencies compared to CNN and Transformer-based methods. but it falls short in refining local interactions.



In this work, we introduce MambaLLIE, a novel framework for enhancing low light images that integrates an implicit Retinex-aware approach within a global-then-local state space model. MambaLLIE not only explores the capabilities of state space models in low light image enhancement but also incorporates a Retinex-aware structure providing both explicit and implicit guidance. Our framework introduces a unique global-then-local state space block, enhancing global long-range degradation modeling and local feature aggregation through an augmented state space. Additionally, we incorporate a Retinex-aware selective kernel mechanism in the enhancement process, enabling adaptive modulation of illumination strength through specific spatial operations.


Our contributions and main findings can be summarized as follows: \text{1)} We introduce a novel global-then-local state space block that integrat a local-enhanced state space module and an implicit Retinex-aware selective kernel module. This design effectively captures intricate global and local dependencies.
\text{2)} We devise an implicit Retinex-aware selective kernel mechanism to guide deeper neural representations, eliminating the need for complex structural design and constraints to estimate physical priors, the prior feature tends to segregate into independent positive and negative illumination components before integrating them, a capability lacking in explicit methods.
\text{3)} Experimental results on benchmark datasets and real-world evaluations consistently demonstrate the superior performance of our proposed method compared to the state-of-the-art approaches.


\section{Related work}

\textbf{Low Light Image Enhancement.}
Nowadays, the exciting deep learning-based methods have mainly been categorized into end-to-end  and Retinex-based methods \cite{LLIEsurvey}. To the best our knowledge, LLNet \cite{LLNet} firstly introduced a deep neural network for low light image enhancement by supervised learning. LightenNet \cite{LightenNet} adopted the CNN for single image contrast enhancement. MBLLEN \cite{MBLLEN} proposed the multi-branch fusion within CNN to extract rich features. Besides, SNR-Net \cite{SNR-Net}, Restormer \cite{Restormer} LLFormer \cite{LLFormer} and \cite{LvZWZYH23} adopted the self-attention mechanism to achieve excellent performance. However, all these end-to-end models mainly depend on the distribution of training dataset and ignore the inherent illumination prior. As contrast, ZeroDCE \cite{ZeroDCE}, RUAS \cite{RUAS}, and subsequent works \cite{SCI,PairLIE,quadprior} represent impressive solutions for image enhancement, as ones precisely using physical priors to enhance the images. 
However, due to the absence of an ideal reference for guidance, these methods usually exhibit a certain gap compared to the supervised learning models. 

As for the supervised Retinex-based models, these methods aim to decompose the image into illumination and reflectance maps, and then enhance the image by optimizing these maps. For instance, Retinex-Net \cite{RetinexNet} divided image enhancement into decomposition, adjustment and reconstruction stages, which providing a good representation of image enhancement process. KinD \cite{KinD} and URetinex-Net \cite{URetinex} further introduced the novel multi-branch and multi-stage frameworks, respectively. However, striking a balance between complexity and efficiency remains challenging for these methods. Recently, RetinexFormer \cite{RetinexFormer} simplified a one-stage Retinex-based low light enhancer with a efficient Transformer. Diff-Retinex \cite{Diff-Retinex} designed a transformer-based decomposition network and adopted generative diffusion networks to reconstruct the results. Overall, they typically applied the Retinex theory in a direct way, which may be limited for low light enhancement problem.

\textbf{Vision State Space Model.} 
State Space Model (SSMs) \cite{LinearStateSpaceLayers,S4,S4D} are burgeoning new sequence models for deep learning, which first swept the natural language processing (NLP) community such as language understanding \cite{S4plus}, content-based reasoning \cite{AugmentedTransformer}. Recently, SSMs have also garnered considerable attention in computer vision (CV) tasks. To our knowledge, S4ND \cite{S4ND} first explored state space mechanism into CV tasks by swapping Conv2D and self-attention layers with S4ND layers in existing models. VMamba \cite{Vmamba} bridged the gap between ordered sequences and non-causal visual images, enabling the extension of vision selective state space model with global receptive fields. Vim \cite{vim} proposed the bidirectional state space modeling with positional awareness, achieved the global visual perception. Furthermore, LocalMamba \cite{LocalMamba} was focused on the local scanning strategy, preservation of local context dependencies. EfficientVMamba \cite{EfficientVMamba} designed a light-weight SSMs with an additional convolution branch to learn both global and local representational features. MambaIR \cite{MambaIR} employed convolution and channel attention to enhance the capabilities of the Mamba. But existing vision state space model does not pay enough attention on capturing local information, as vanilla SSMs are designed for long sequence and the invariant of local vision data is not taken into account in the existing vision state space models.  

\section{Methodology}

This work aims to introduce a novel implicit Retinex-aware low light enhancer with global-then-local state space. In this section, we revisit the Retinex theory and the state space model, offering a concise overview. Following that, the details of our proposed MambaLLIE are introduced.

\subsection{Preliminaries}

\textbf{Retinex Theory.}
The ideal Retinex theory \cite{land1971lightness} for low light enhancement assumes that the captured images can be decomposed into reflectance and illumination maps. Following \cite{SCI,DI-Retinex}, explicit Retinex-based methods emphasize estimating either an illumination map while regarding the reflectance map as the enhanced result, or estimating concrete reflectance and illumination maps and then restoring the well-exposed images. Specifically, given a low light image ${\bf{L}} \in \mathbb{R}^{H \times W \times 3}$, where $H$ and $W$ represent height and width respectively, the derived maps can be denoted as:
\begin{align}
    \bf{L} = \bf{R} \cdot \bf{I},\quad 
\bf{N} = {\bf{L} \mathord{\left/
 {\vphantom {L {\tilde I}}} \right.
 \kern-\nulldelimiterspace} {\tilde I}},\quad 
\bf{N} = \bf{\tilde R} \cdot \bf{\tilde I}.
\end{align}
where $\cdot$ denotes the element-wise multiplication,  ${\bf{R}} \in \mathbb{R}^{H \times W \times 3}$ denotes reflectance map, a static property of captured objects; ${\bf{I}} \in \mathbb{R}^{H \times W}$ denotes illumination map; ${\bf{N}} \in \mathbb{R}^{H \times W \times 3}$ denotes normal images; ${{\bf{\tilde R}}}$, ${{\bf{\tilde I}} \in \mathbb{R}^{H \times W \times 3}}$ denotes the estimated reflectance and illumination maps, respectively. 

Consequently, the former assumption ignore the noise and artifacts resulting from sensor degradation in the captured images, while pixel-wise adjustments for the illumination is inadequate.  The latter aims to restore the reflectance and illumination maps to enhance the images. However, this requires the design of multiple branches and constraints to guide the training \cite{KinD}.

\textbf{State Space Model.}
The SSMs, such as structured state space sequence models (S4) \cite{S4} and Mamba \cite{Mamba}, can be regarded as the continuous linear time-invariant (LTI) systems \cite{LTI}. Given an one-dimension sequence $x\left( t \right) \in \mathbb{R}$, it projects into a new one-dimension sequence $y\left( t \right) \in \mathbb{R}$ through the hidden state $h\left( t \right) \in \mathbb{R}^{m}$, the whole system can be defined as a linear ordinary differential equation (ODE):
\begin{align}
\begin{aligned}
h'(t) &= {\bf{A}}h(t) + {\bf{B}}x(t), \\
y(t) &= {\bf{C}}h(t) + {\bf{D}}x(t).
\end{aligned}
\end{align}
where $m$ denotes the state size, ${\bf{{A}}} \in \mathbb{R}^{m \times m}$, ${\bf{{B}}} \in \mathbb{R}^{m \times 1}$, ${\bf{{C}}} \in \mathbb{R}^{1 \times m}$ and ${\bf{{D}}} \in \mathbb{R}$ denotes state, input projection, output projection, and feedthrough parameters.


As the raw state-space models are continuous, the systems adopt the discrete versions before feeding the computer, in which the zero-order hold (ZOH) is used to transform the continuous parameters ${\bf{ {A}}}$ and ${\bf{{B}}}$ to discrete parameters ${\bf{\bar {A}}}$ and ${\bf{\bar {B}}}$ as follows
\begin{align}
\begin{array}{l}
{\bf{\bar {A}}} = \exp \left( {\Delta {\bf{{A}}}} \right),\quad 
{\bf{\bar B}} = {\left( {\Delta {\bf{{A}}}} \right)^{ - 1}}\left( {\exp \left( {\Delta {\bf{{A}}}} \right) - {\bf{I}}} \right) \cdot \Delta {\bf{B}}.
\end{array}
\end{align}
where $\Delta$ denotes a step size. Overall, the discretized version can be rewritten as:
\begin{align}
\begin{array}{l}
{h_t} = {\bf{\bar {A}}}{h_{t - 1}} + {\bf{\bar B}}{x_t}, \quad  {y_t} = {\bf{C}}{h_t} + {\bf{D}}{x_t}.
\end{array}
\end{align}

However, the current system remains static for varying inputs. To address this limitation, Mamba \cite{Mamba} introduces selective state space models, allowing parameters to adapt with the input, thereby enhancing selective information processing across sequences. This parameter selection mechanism can be expressed as:
\begin{equation}
\begin{aligned}
\overline{{\bf{B}}} & =f_{{\bf{B}}}({x_t}), \quad 
\overline{{\bf{C}}} & =f_{{\bf{C}}}({x_t}), \quad 
\Delta & =\vartheta_{{\bf{A}}}\left({{\bf{P}}}+f_{{\bf{A}}}({x_t})\right).
\end{aligned}
\end{equation}
where $f_{{\bf{B}}}({x_t})$, $f_{{\bf{C}}}({x_t})$ and $f_{{\bf{A}}}({x_t})$ are linear functions that broadens feature to the hidden state dimensions. As SSMs are tailored for long sequences, it is limited in capturing complicated local information. For visual data, VMamba \cite{Vmamba} and Vim \cite{vim} proposed the specific location-aware scan strategies to maintains the integrity of 2D image structures. However, the specific directed sequences overlook the vision information of pixels neighborhood structure. 
Inspired by \cite{AugmentedTransformer}, we explore a global-then-local state space, which receives the global perception before the details, supplementing the lack of local information.

\begin{figure}
  \centering
  \includegraphics[width=\textwidth]{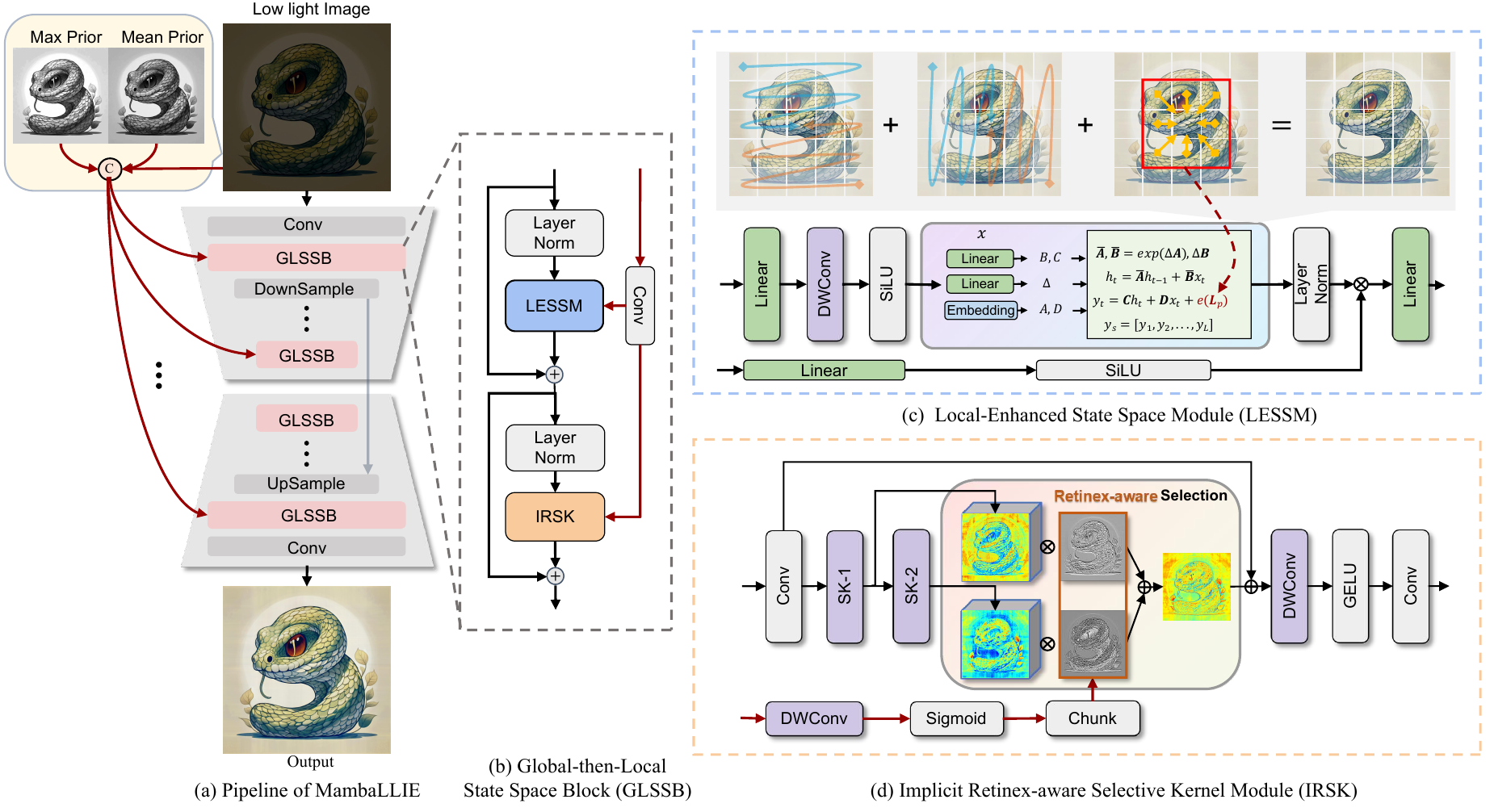}
  \caption{The overall pipeline of the proposed MambaLLIE. Our Global-then-Local State Space Block (GLSSB) integrates Local-enhanced state space module (LESSM) and implicit Retinex-aware selective kernel module (IRSK) with layer normalization as its core.}
  \label{fig:framework}
  \vspace{-0.2in}
\end{figure}

\subsection{Overall Pipeline}

We first present the overall pipeline of our MambaLLIE, an U-shaped architecture as shown in Fig. \ref{fig:framework}(a), which includes encode and decode parts with the convolutional downsampling and upsampling layers. The encoder features are concatenated with the decoder features via skip connections. Next, We propose a global-then-local state space block (GLSSB) as the basic core of MambaLLIE, the max and mean priors concatenated with low light image are projected into GLSSB by convolutional layers as augmented input. Therein, GLSSB is composed of the local-enhanced state space module (LESSM) and the implicit Retinex-aware selective kernel module (IRSK), interleaved with layer norm layer. 

Specifically, given a low light image ${\bf{L}}\in \mathbb{R}^{H \times W \times 3}$, we employ a $3 \times 3$ convolution layer to project the neural features ${\bf{F}} \in \mathbb{R}^{H \times W \times C}$ from input feature space, and then project features into each GLSSB, which will be described in Section 3.3. Besides, IRSK integrates original input, maximum prior ${\bf{{L_{\max }}}} \in \mathbb{R}^{H \times W}$ and mean prior ${\bf{{L}}}_{mean} \in \mathbb{R}^{H \times W}$ as augmented input ${\bf{{L}}}_p  \in \mathbb{R}^{H \times W \times 5}$, 
\begin{align}
  \mathbf{L}_p = \mathrm{Concat}\left( \mathbf{L}, \mathrm{mean}(\mathbf{L}), \max(\mathbf{L}) \right).
\end{align}
We first define ${{\bf{F}}_g}$ is the output of GLSSB. Subsequently, the downsampling layer and following GLSSB achieve the feature extraction to acquire the deep feature, which can be denoted as ${{\bf{F}}_g} \in {\mathbb{R}^{\frac{H}{{{2^i}}} \times \frac{W}{{{2^i}}} \times {2^i}C}}$, where $i=0,1,2$. Moreover, the feature is later concatenated with the upsampling layer with a symmetrical structure. Finally, using a $3 \times 3$ convolution layer projects into ${{\bf{F}}}_{out} \in \mathbb{R}^{H \times W \times 3}$ and the enhanced image can be expressed as $\mathbf{N} = {{\bf{F}}_{out}} + \mathbf{L}$.

\subsection{Global-then-Local State Space Block}

As illustrated in Fig. \ref{fig:framework}(b), GLSSB follows the LayerNorm, LESSM, LayerNorm and IRSK flow, motivated by Transformer \cite{attention} and Mamba\cite{Mamba} usage of similar structures in a basic block. Given the input feature, it first undergoes the LayerNorm and LESSM to capture the local-enhanced global information. the above process can be denoted as:
\begin{align}
\mathbf{M} = \mathrm{LESSM}\left( \mathrm{LN}\left( \mathbf{F}_g^{i-1} \right) \right) + \mathbf{F}_g^{i-1}.
\end{align}
And then, another LayerNorm and our proposed IRSK are used for Retinex-aware guidence. The above process can be formulated as:
\begin{align}
\mathbf{F}_g^i = \mathrm{IRSK}\left( \mathrm{LN}\left( \mathbf{M} \right) \right) + \mathbf{M}.
\end{align}

Overall, at the prior module of GLSSB, we capture global dependencies using a local-enhanced SSM.  Because the SSM is better at learning global information, the subsequent module aims to handle more refined and complicated local dependencies.

\textbf{Local-Enhanced State Space Module}. Existing state space models \cite{H3,S4,Mamba} excels at capturing the causal processing of input data in long range dependencies. However, the unidirectional scan manner encounters difficulties in vision data to modeling non-causal relationships. To accommodate vision data, \cite{vim,Vmamba,EfficientVMamba} process the input data from different 2D scan directions. However, these methods ignore the local invariants of vision data. In other word, the fixed scanning methods widen the distance between neighborhood data and snarl the causal relationships.

The most SSMs \cite{Vmamba} can be regarded as the continuous linear time-invarian systems, we further introduce the a ${e}\left({{\bf{L_p}}}\right)$ augmented local bias, enhancing the original SSMs by preserving local 2D dependency as shown in Fig. \ref{fig:framework}(c). Following \cite{yang1997observer,IBRIR2008579}, we propose a global-then-local state space:
\begin{align}
\begin{array}{l}
{h_t} = {\bf{\bar {A}}}{h_{t - 1}} + {\bf{\bar B}}{x_t},\\
{y_t} = {\bf{C}}{h_t} + {\bf{D}}{x_t} + {e}\left({{\bf{L_p}}}\right).
\end{array}
\end{align}
where ${e}\left({{\bf{L_p}}}\right)$ is independent of the hidden state space. Hence, the model can be
computed in a simple way, given a feature ${\bf{F}} \in \mathbb{R}^{H \times W \times 5}$ and illumination feature ${\bf{{L_{p }}}} \in \mathbb{R}^{H \times W \times C}$, we adopts the Layernorm followed by our proposed LESSM to integrate the spatial long-term dependency. Following \cite{Mamba}, the input feature are chunk into ${\bf{{\tilde F}_1}}$ and ${\bf{{\tilde F}_2}}$ in two branches. The first branch projects the feature into a linear layer, followed by a depth-wise convolution, SiLU activation function, accompanied by our proposed augmented local bias and Layernorm. In the second branch, the features is also projected to a linear layer followed by the SiLU activation function. Finally, features from the two branches are aggregated with the element-wise product and then are projected back to input feature space by linear layer. The entire process can be delineated as
\begin{equation}
\begin{aligned}
& \mathbf{\widetilde{F}}_1 = \operatorname{LN}\left( \operatorname{2DSSM}\left(\operatorname{SiLU}\left(\operatorname{DWConv}\left(\operatorname{Linear}\left(\mathbf{F}_1\right)\right)\right)\right) + \operatorname{Conv}(\mathbf{L}_p)\right) \\
& \widetilde{\mathbf{F}}_2 = \operatorname{SiLU}\left(\operatorname{Linear}\left(\mathbf{F}_2\right)\right)\\
& \widetilde{\mathbf{F}}_{\text{out}} = \operatorname{Linear}\left(\widetilde{\mathbf{F}}_1 \odot \widetilde{\mathbf{F}}_2\right)
\end{aligned}
\end{equation}

\textbf{Implicit Retinex-Aware Selective Kernel Module.}
We further employ a implicit Retinex-aware selective kernel network to enhance the capabilities of the information integration ability. IRSK constructs a sequence of depth-wise convolutions with an alterable kernel to select the feature with different receptive field, using a spatial selection mechanism by illumination prior. Inspired by LSKNet \cite{LSKNet}, for each of the feature maps from different selective kernel, a Sigmoid activation function is applied to obtain the individual illumination maps from illumination prior. Fig. \ref{fig:framework}(d) shows a detailed conceptual illustration of IRSK module where we intuitively demonstrate how the implicit Retinex-aware module works. The above process can be formulated as:

\begin{equation}
\widetilde{\mathbf{F}}_k = \widetilde{\mathbf{F}}_{\text{out}}, \quad \widetilde{\mathbf{F}}_{k+1} = f_{\text{DWconv}}^{k}\left(\widetilde{\mathbf{F}}_k\right).
\end{equation}

The output of the Retinex-aware maps are concatenated with the input features via residual
connections, followed by a depth-wise convolution, GELU activation function and convolution layer.
\begin{equation}
\{ \mathbf{S}_1, \mathbf{S}_2 \} = \operatorname{Chunk}\left( \operatorname{Sigmoid}\left( \operatorname{Conv} \mathbf(\mathbf{L}_p)  \right) \right)
\end{equation}
\begin{equation}
\mathbf{F}_g = \operatorname{Conv} \bigl( \operatorname{GELU} \bigl( \operatorname{DWConv} \bigl( \sum_{k = 1}^K \widetilde{\mathbf{F}}_k \mathbf{S}_k + \widetilde{\mathbf{F}}_{\text{out}} \bigr) \bigr) \bigr)
\end{equation}

\section{Experiments}
\label{Experiments}

\subsection{Benchmark Datasets and Implementation Details}
\label{ImplementationDetails}

\textbf{Datasets.} We employ five paired low light image datasets for evaluation, including LOL-V2-real \cite{LOLV2}, LOL-v2-syn \cite{LOLV2}, SMID \cite{SMID}, SDSD-indoor \cite{SDSD} and SDSD-outdoor \cite{SDSD} datasets. Therein, LOL-V2-real contains 689 low-normal light paired images for training and 100 pairs for testing; LOL-V2-syn includes 900 paired images for training and the 100 pairs for testing; Besides, SMID is composed of the 15763 short-long exposure paired images for training and the remaining images for testing; SDSD-indoor and SDSD-outdoor are all subsets of SDSD dataset (the static version), which extract the paired images from 62 and 116 pairs for training, and the left 6 and 10 pairs for testing.

\textbf{Implementation Details}. We implement MambaLLIE in PyTorch \cite{torch} on a server with the 4090GPUs. Random cropping the image pairs into $128 \times 128$
patches as training samples, data augmentation is performed on the training samples such as rotation and flipping. The batch size is 8. In terms of optimization procedure, Adam \cite{Adam} is adopted as the optimizer with ${\beta _1} = 0.9$ and ${\beta _2} = 0.999$; The training iterations is set to $1.5 \times 10^5$. The initial e learning rate is set to $2 \times 10^{-4}$ and steadily decreased by by the cosine annealing scheme. The loss criterion is mean absolute error (MAE), thus peak signal-to-noise ratio (PSNR) and structural similarity (SSIM) \cite{SSIM} is selected as the evaluation metrics for the paired datasets.

\begin{table}[t]
\caption{Quantitative comparisons on LOL-V2-real, LOL-V2-syn, SMID, SDSD-indoor and SDSD-outdoor datasets. The best result is in {\color[HTML]{FF0000} red} color while the second best result is in {\color[HTML]{4874CB} blue} color.}
\label{Quantitative comparisons}
\centering
\resizebox{\textwidth}{!}{%
\begin{tabular}{c|c|cc|cc|cc|cc|cc|cc}
\hline
                          &                        & \multicolumn{2}{c|}{LOL-V2-real}                                              & \multicolumn{2}{c|}{LOL-V2-syn}                                               & \multicolumn{2}{c|}{SMID}                                                     & \multicolumn{2}{c|}{SDSD-indoor}                                              & \multicolumn{2}{c|}{SDSD-outdoor}                                             & \multicolumn{2}{c}{Complexity}                                                                             \\ \cline{3-14} 
\multirow{-2}{*}{Methods} & \multirow{-2}{*}{Ref.} & PSNR                                  & SSIM                                  & PSNR                                  & SSIM                                  & PSNR                                  & SSIM                                  & PSNR                                  & SSIM                                  & PSNR                                  & SSIM                                  & FLOPS& Param \\ \hline
RetinexNet                & BMVC 2018              & 15.47                                 & 0.567                                 & 17.13                                 & 0.798                                 & 22.83                                 & 0.684                                 & 20.84                                 & 0.617                                 & 20.96                                 & 0.629                                 & 587.47                                              & 0.84                                                 \\
DeepUPE                   & CVPR 2019              & 13.27                                 & 0.452                                 & 15.08                                 & 0.623                                 & 23.91                                 & 0.690                                 & 21.70                                 & 0.662                                 & 21.94                                 & 0.698                                 & 21.10                                               & 1.02                                                 \\
SID                       & ICCV 2019              & 13.24                                 & 0.442                                 & 15.04                                 & 0.610                                 & 24.78                                 & 0.718                                 & 23.29                                 & 0.703                                 & 24.90                                 & 0.693                                 & 13.73                                               & 7.76                                                 \\
KinD                      & MM 2019                & 14.74                                 & 0.641                                 & 13.29                                 & 0.578                                 & 22.18                                 & 0.634                                 & 21.95                                 & 0.672                                 & 21.97                                 & 0.654                                 & 34.99                                               & 8.02                                                 \\
MIRNet                    & ECCV 2020              & 20.02                                 & 0.820                                 & 21.94                                 & 0.876                                 & 25.66                                 & 0.762                                 & 24.38                                 & 0.864                                 & 27.13                                 & 0.837                                 & 785.00                                              & 31.76                                                \\
EnGAN                     & TIP 2021               & 18.23                                 & 0.617                                 & 16.57                                 & 0.734                                 & 22.62                                 & 0.718                                 & 23.29                                 & 0.703                                 & 24.90                                 & 0.693                                 & 61.01                                               & 114.35                                               \\
Restormer                 & CVPR 2022              & 19.94                                 & 0.827                                 & 21.41                                 & 0.830                                 & 26.97                                 & 0.758                                 & 25.67                                 & 0.827                                 & 24.79                                 & 0.802                                 & 144.25                                              & 26.13                                                \\
SNR-Net                   & CVPR 2022              & 21.48                                 & {\color[HTML]{FF0000} 0.849} & 24.14                                 & 0.928                                 & 28.49                                 & 0.805                                 & 29.44                                 & 0.894                                 & 28.66                                 & 0.866                                 & 26.35                                               & 4.01                                                 \\
QuadPrior                 & CVPR 2024              & 20.48                                 & 0.811                                 & 16.11                                 & 0.758                                 & 15.50                                 & 0.604                                 & 22.22                                 & 0.783                                 & 18.26                                 & 0.662                                 & /                                                   & /                                                    \\
MambaIR                   & Arxiv 2024             & 21.25                                 & 0.831                                 & 25.55                                 & 0.929                                 & 27.07                                 & 0.774                                 & 28.97                                 & 0.884                                 & 29.75                                 & 0.861                                 & 60.66                                               & 4.30                                                 \\
RetinexFormer             & ICCV 2023              & {\color[HTML]{4874CB} 22.80}          & 0.840                                 & {\color[HTML]{4874CB} 25.67}          & {\color[HTML]{4874CB} 0.930}          & {\color[HTML]{4874CB} 29.15}          & {\color[HTML]{4874CB} 0.815}          & {\color[HTML]{4874CB} 29.77}          & {\color[HTML]{4874CB} 0.896}          & {\color[HTML]{4874CB} 29.84}          & {\color[HTML]{FF0000} 0.877} & 15.57                                               & 1.61                                                 \\
MambaLLIE                 & /                      & {\color[HTML]{F5222D} 22.95} & {\color[HTML]{4874CB} 0.847}          & {\color[HTML]{F5222D} 25.87} & {\color[HTML]{F5222D} 0.940} & {\color[HTML]{F5222D} 29.26} & {\color[HTML]{FF0000} 0.818} & {\color[HTML]{F5222D} 30.12} & {\color[HTML]{F5222D} 0.900} & {\color[HTML]{F5222D} 30.00} & {\color[HTML]{4874CB} 0.869}          & 20.85                                               & 2.28                                                 \\ \hline
\end{tabular}
}
\end{table}

\begin{figure}[t]
  \centering
  \includegraphics[width=0.97\textwidth]{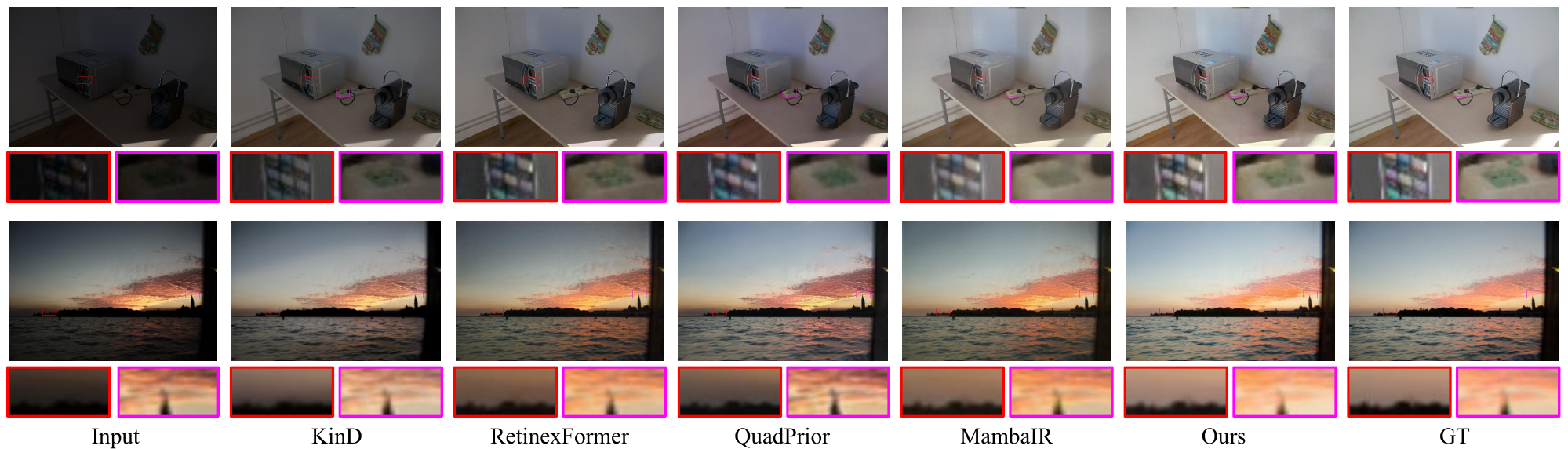}
  \vspace{-5pt}
  \caption{Qualitative comparison with previous methods on LOL-V2-real and LOL-V2-syn datasets. Our MambaLLIE effectively enhances the illumination and preserves the color.}
  \label{fig:lol_res}
  \vspace{-10pt}
\end{figure}

\begin{figure}
  \centering
  \includegraphics[width=0.97\textwidth]{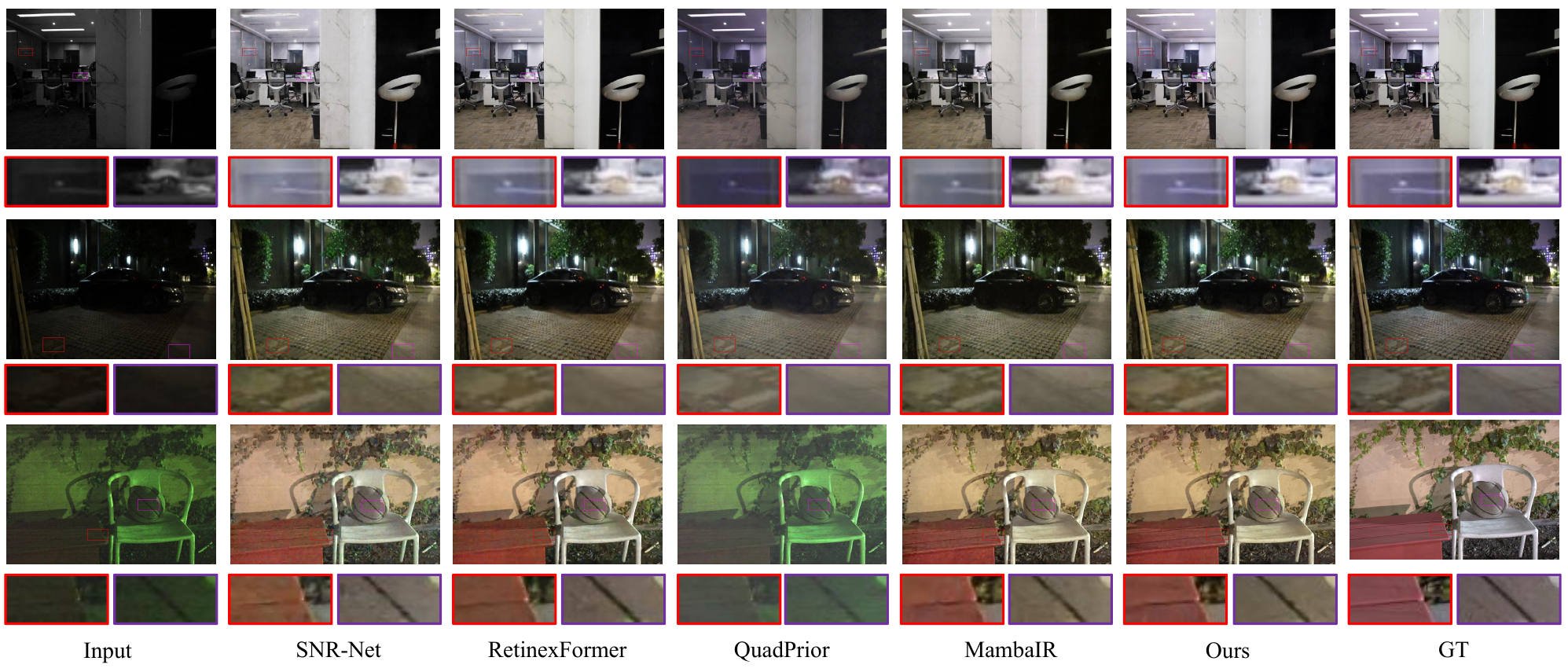}
  \caption{Qualitative comparison with previous methods on SMID, SDSD-indoor and SDSD-outdoor datasets. Our MambaLLIE restore the texture and color under challenging degradation, such as the wooden bench and reflective glass.}
  \label{fig:SDSD_res}
  \vspace{-0.2in}
\end{figure}

\subsection{Main Results on Benchmarks.}
\label{benchmarks}

\textbf{Quantitative Comparison.} As shown in Tab. \ref{Quantitative comparisons}, we evaluated the performance of our MambaLLIE against 11 SOTA image enhancement methods, including RetinexNet \cite{RetinexNet}, DeepUPE \cite{DeepUPE}, SID \cite{SMID}, KinD \cite{KinD}, MIRNet \cite{MIRNet}, EnGan \cite{EnGan}, Restormer \cite{Restormer}, SNR-Net \cite{SNR-Net}, QuadPrior \cite{quadprior}, MambaIR \cite{MambaIR} and Retinexformer \cite{RetinexFormer}. Our MambaLLIE demonstrates superior performance than SOTA methods on the adopted benchmark datasets in terms of PSNR and SSIM, while achieves comparable results of SSIM with the SOTA methods in LOL-V2-real and SDSD-outdoor. 
Therein, when the parameters are roughly similar, our MambaLLIE achieves an average improvement of 0.2 dB on benchmark datasets compared to the previous Transformer based SOTA method, \textit{i.e.} RetinexFormer. Compared with the earlier Transformer based SNR-Net, MambaLLIE outperforms it by average 1dB PSNR on the all datasets. When compared to the MambaIR, MambaLLIE achieves 1.70, 0.32, 2.19, 1.15 and 0.25 dB PSNR improvements on the adopted datasets, respectively. Besides, Our MambLLIE gains the improvements over 7 dB on all datasets than traditional Retinex-based models, such as RetinexNet, DeepUPE and KinD. 

\textbf{Qualitative Comparison.} Figs. \ref{fig:lol_res} \& \ref{fig:SDSD_res} report the vision results for comparing our method with latest the SOTA methods and traditional Retinex-based models. Existing methods suffer from insufficient illumination and fail to restore the details as shown in Fig. \ref{fig:lol_res}. As we can see, color distortion and image degradation also affect the enhanced results of previous methods in Fig. \ref{fig:SDSD_res}, yet our MambaLLIE not only enhances brightness but also faithfully preserves colors with reference to ground truth images, all while restoring the details.
 
\subsection{Real World Experimental Evaluation}
\label{exdark}

Enhancing low light images in real-world scenarios is exceptionally challenging because not only can provide benefits for downstream tasks, such as dark object detection, but must the enhanced images be pleasing to the human perception. 

\begin{figure}[t]
  \centering
  \includegraphics[width=0.97\textwidth]{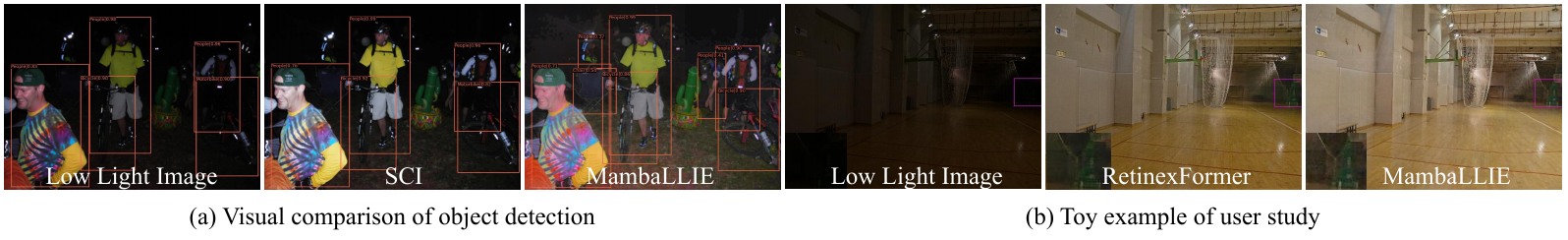}
  \caption{Vision comparison of our MambaLLIE with recent SOTA methods.(a) Qualitative comparison on object detection, (b) A toy example of user study.}
  \label{fig:object}
  \vspace{-0.1in}
\end{figure}

\begin{table}[h]
\centering
\caption{Low light object detection results on the ExDark dataset. The best result is in {\color[HTML]{FF0000} red} color while the second best result is in {\color[HTML]{4874CB} blue} color.}
\label{table:object}
\resizebox{\textwidth}{!}{%
\begin{tabular}{c|ccccccccccccc}
\hline
Methods       & Bicycle                      & Boat                         & Bottle                       & Bus                          & Car                          & Cat                          & Chair                        & Cup                          & Dog                          & Motor                        & People                       & Table                        & Mean                         \\ \hline
RetinexNet    & 0.790                        & 0.741                        & 0.743                        & 0.908                        & 0.820                        & 0.665                        & 0.651                        & 0.750                        & 0.721                        & 0.703                        & 0.784                        & 0.556                        & 0.736                        \\
EnGAN         & 0.733                        & 0.710                        & 0.687                        & 0.892                        & 0.786                        & 0.675                        & 0.656                        & 0.650                        & 0.741                        & 0.657                        & 0.731                        & 0.528                        & 0.704                        \\
KinD          & 0.800                        & 0.721                        & {\color[HTML]{FF0000} 0.788} & 0.919                        & 0.822                        & {\color[HTML]{FF0000} 0.718} & 0.672                        & 0.771                        & 0.775                        & 0.736                        & 0.803                        & 0.555                        & 0.757                        \\
ZeroDCE       & 0.806                        & 0.750                        & 0.762                        & 0.914                        & 0.837                        & 0.681                        & 0.677                        & 0.769                        & {\color[HTML]{FF0000} 0.788} & 0.728                        & 0.801                        & 0.535                        & 0.754                        \\
SCI           & {\color[HTML]{FF0000} 0.821} & 0.742                        & 0.749                        & 0.916                        & {\color[HTML]{FF0000} 0.846} & 0.695                        & {\color[HTML]{4874CB} 0.690} & {\color[HTML]{4874CB} 0.784} & 0.756                        & {\color[HTML]{FF0000} 0.758} & 0.810                        & 0.555                        & {\color[HTML]{FF0000} 0.760} \\
SNR-Net       & 0.802                        & 0.721                        & 0.750                        & {\color[HTML]{FF0000} 0.932} & 0.840                        & 0.694                        & 0.677                        & 0.758                        & 0.763                        & {\color[HTML]{4874CB} 0.755} & 0.789                        & 0.559                        & 0.753                        \\
Retinexformer & {\color[HTML]{4874CB} 0.809} & {\color[HTML]{FF0000} 0.769} & 0.753                        & 0.914                        & 0.814                        & 0.688                        & 0.689                        & 0.763                        & 0.766                        & 0.769                        &  0.805 & 0.543                        & 0.757                        \\
MambaIR       & 0.803                        & 0.763                        & 0.752                        & 0.903                        & 0.830                        & 0.687                        & 0.684                        & 0.761                        & 0.721                        & 0.738                        & {\color[HTML]{FF0000} 0.813}                        & 0.556                        & 0.751                        \\
MambaLLIE     & 0.802                        & {\color[HTML]{0070C0} 0.764} & {\color[HTML]{4874CB} 0.779} & {\color[HTML]{4874CB} 0.926} & {\color[HTML]{FF0000} 0.846} & {\color[HTML]{4874CB} 0.701} & {\color[HTML]{FF0000} 0.692} & {\color[HTML]{FF0000} 0.800} & {\color[HTML]{4874CB} 0.781} & 0.751                        & {\color[HTML]{4874CB}0.812} & {\color[HTML]{FF0000} 0.560} & {\color[HTML]{FF0000} 0.768} \\ \hline
\end{tabular}}
\end{table}

\textbf{Low Light Object Detection.}
We utilized ExDark dataset \cite{Exdark} to compare the enhancement of preprocessing methods for high-level vision tasks. There are 7363 challenging low light images annotated with 12 bounding box classes, of which 5,890 for training and 1,473 for testing. Note that all supervised methods were pretrained on the LOL-V2-syn dataset, the low light image underwent different enhancement methods and then finetuned YOLOv3 \cite{yolov3} as the object detector.

As shown in Tab. \ref{table:object}, our methods achieved the best average result compared with all adopted models, and yielded the best results on Car, Chair, Cup and Table classes. Fig. \ref{fig:object}(a) further reported the visual comparison, compared with suboptimal preprocessing method SCI, detector through our MambaLLIE can detecte the objects in extreme dark regions including two persons and a chair, while other methods failed.

\textbf{User Study.}
We conducted a user study to evaluate the human visual perception quality of the enhanced results in challenging scenarios. Due to the lack of the ideal reference for training, we selected the pretrained model from the benchmarks to enhance the photos. There are 7 random selected low light images from the benchmarks and ExDark datasets under different lighting conditions. Human perception primarily focuses on the presence of global visual effect, local detail, color distortion (noise), which significantly reflect the quality of the enhanced images. Thus, We assigned ratings on a scale of 1 (worst) to 5 (best), evaluating the quality of the enhancements in terms of overall rating, local detail and color distortion(noise), respectively. Overall, 70 participants were invited to assess the visual quality. The average scores are reported in Tab. \ref{table:userstudy}, our MambaLLIE achieves the best score in the involved voting aspects. Fig. \ref{fig:object}(b) shows the toy example of user study, which display the input and the random enhanced results and local details by different algorithms.

\begin{table}[h]
\centering
\caption{User study on the challenging low light image enhancement.}
\label{table:userstudy}
\resizebox{\textwidth}{!}{%
\begin{tabular}{c|cccccccc}
\hline
Methods             & RetinexNet & EnGAN & SCI   & QuadPrior & SNR-Net & Retinexformer & MambaIR & MambaLLIE                    \\ \hline
Overall Rating      & 3.093      & 3.314 & 3.943 & 3.014     & 3.821   & 4.100         & 3.857   & \textbf{4.243} \\
Local Detail        & 2.871      & 3.143 & 3.686 & 3.129     & 3.779   & 3.950         & 3.629   & \textbf{4.129} \\
Color distortion(noise) & 2.914      & 3.164 & 3.776 & 2.929     & 3.657   & 3.971         & 3.750   & \textbf{4.100} \\ \hline
\end{tabular}}
\end{table}




\begin{table}[t]
  \vspace{-0.1in}
\centering
\begin{minipage}[t]{0.44\textwidth}
\centering
\caption{Effects of design choices.}
\label{table:abl_baseline}
\resizebox{\textwidth}{!}{%
\begin{tabular}{c|cc|cc}
\hline
Methods           & Params & FLOPS & PSNR           & SSIM           \\ \hline
Baseline-1 & 2.14      & 18.39    & 28.87          & 0.865          \\
Baseline-2 & 2.14      & 18.39    & 29.12          & 0.862          \\
Ours w/o LESSM       & 2.26      & 20.64    & 29.83          & 0.889          \\
Ours w/o IRSK      & 2.19      & 19.94    & 29.20          & 0.887          \\
Ours              & 2.28      & 20.85    & \textbf{30.12} & \textbf{0.900} \\ \hline
\end{tabular}}
\end{minipage}%
\hspace{0.04\textwidth}
\begin{minipage}[t]{0.44\textwidth}
\centering
\caption{Effects of different selective kernel.}
\label{table:sk}
\resizebox{\textwidth}{!}{%
\begin{tabular}{c|cc|cc}
\hline
Kernel Sizes        & Params     & FLOPS      & PSNR           & SSIM           \\ \hline
3*3                 & 2.25          & 20.47          & 29.55          & 0.899          \\
5*5                 & 2.31          & 21.23          & 29.48          & 0.896          \\
5*7                 & 2.35          & 21.79          & 28.88          & 0.892          \\
5*3                 & 2.38          & 20.85          & 29.31          & 0.892          \\
3*5 (Ours)          & 2.28          & 20.85          & \textbf{30.12} & \textbf{0.900} \\ \hline
\end{tabular}}
\end{minipage}
\vspace{-0.1in}
\end{table}

\subsection{Ablation Study}
\textbf{Implicit Retinex-Aware Framework.}
We compare the improvement of using a implicit Retinex-aware model with the end-to-end and explicit Retinex-aware models. Specifically, Baseline-1 is a simple variant of our MambnaLLIE by removing Retinex-aware guidence, namely estimates enhanced result directly from input without any prior. Baseline-2 is designed to estimate the illumination map and then light up the low light image by element-wise multiplication. Tab. \ref{table:abl_baseline} reveals our implicit Retinex-Aware framework significantly outperforms Baseline-1 with the improvement of 1.25 dB in PSNR, while achieving a PSNR enhancement of 1.00 dB compared to Baseline-2.

\textbf{Global-then-Local State Space.}
As the core component, our GLSSB comprises the LESSM and IRSK. We demonstrate the effect of each component through ablation study. The results, presented at the bottom of Tab. \ref{table:abl_baseline}, indicate that our LESSM achieves improvements of 0.33 dB and 0.08 dB in PSNR compared to Baseline-1 and Baseline-2, respectively, which utilize vanilla state space blocks. Additionally, our IRSK produces PSNR enhancements of 0.96, 0.74, and 0.63 dB compared to Baselines and when only applying LESSM. Our full version indicating that although LESSM improves the vanilla SSM with local enhanced modeling ability, IRSK should be considered for further improvements, when GLSSB integrats LESSM and IRSK, our MambaLLIE achieves the highest PSNR and SSIM.

\begin{figure}[t]
  \centering
  \includegraphics[width=0.90\textwidth]{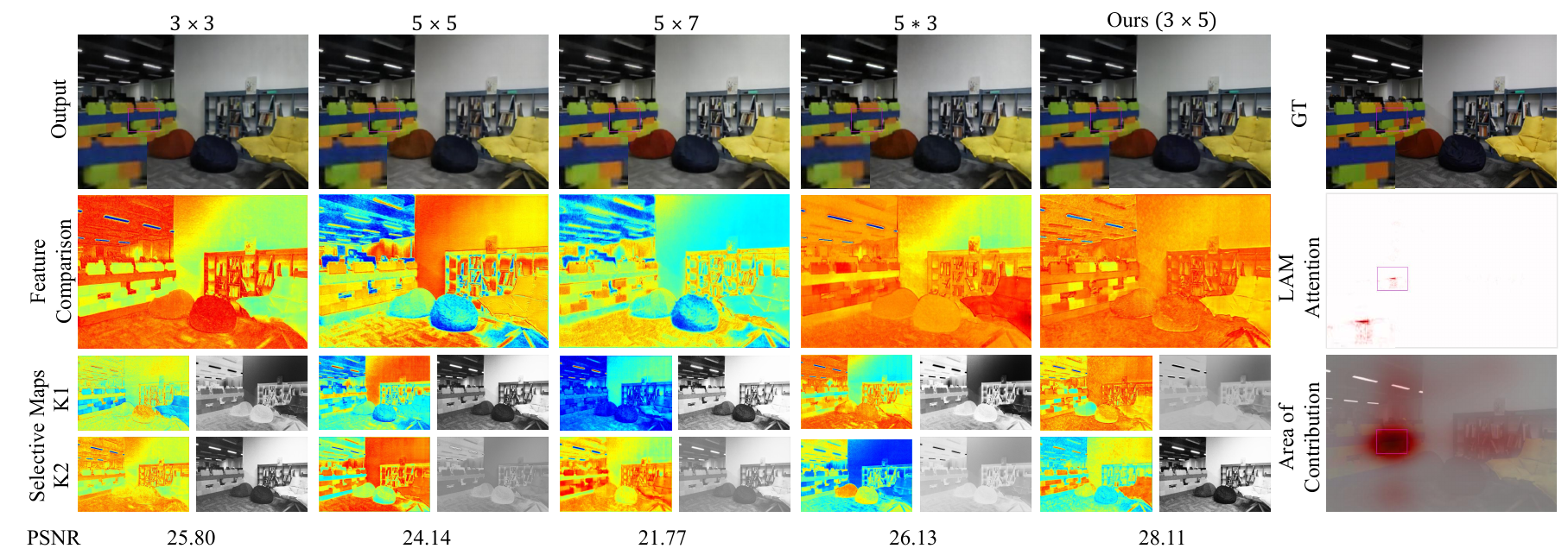}
  \caption{The details of selective kernel behaviour, the LAM visualization \cite{LAM} demonstrates influence of similar local information is higher than that of global dependence, our local-enhanced strategy underscores the feature. Besides, the larger receptive fields can provide globally consistent results. }
  \label{fig:feature}
  \vspace{-0.1in}
\end{figure}

\textbf{Selective Kernel Behaviour.}
We further investigate the kernel selection behaviour in our MambaLLIE as shown in Fig. \ref{fig:feature}. We find the implicit Retinex-aware selection pattern tend to learn two independent positive and negative illumination, resulting in complementary features. Compared with explicit Retinex-based methods, our IRSK can guide from a flexible deeper neural representation. The quantitative results are reported in Tab. \ref{table:sk}. Different with LSKNet \cite{LSKNet}, we put small kernels in front and larger kernels in higher levels. This is because object detection needs larger receptive field, thus adopts a sequence of depth-wise convolutions with growing kernel and increasing dilation, while has to introduce a lots of padding. But image enhancement may suffers from padding operation at the edge of the image, especially upsampling further expands the padding values. Thus, the the former small kernels can quickly focus on local information and the the latter kernels contain larger receptive fields for better feature fusion. 

\section{Limitation and Discussion}
\label{Limitation}

We adopt an implicit Retinex-Aware guidance within a global-then-local state space framework to address global insufficient illumination and local degradation for low light enhancement. However, our approach has several limitations. 1) Unlike end-to-end methods, our technique requires the design of a reasonable illumination prior, which relies on prior experience.
2) Most existing enhancement models, including ours, primarily focus on mean square error and use PSNR and SSIM to evaluate image quality. To mitigate inherent biases in these metrics, we conducted additional real-world experimental evaluations to reconcile the bias and further validate the effectiveness of our approach. 


\section{Conclusion}

 In this paper, we introduced a novel state space-based model, MambaLLIE. Our proposed core of GLSSB effectively combines global and local information by implicit Retinex-aware selective kernel into global-then-local state space. Extensive experiments on benchmarks, low light object detection and user study demonstrate that our framework consistently achieves the best performance. Our future work is to address the dual challenges of local redundancy and global dependencies in low light video enhancement via efficient state space modeling.

\small{\bibliographystyle{plain}
\bibliography{egbib}}
\newpage
\appendix

\section{Broader Impact.}
\label{broader_impact}
Low light image enhancement is the classical task that improves the quality of degraded images, exhibiting the promising value of research and application. Our proposed global-then-local state space enhances the feature extraction ability by integrating implicit Retinex-aware strategy. We believe our method has the potential to advance other low-level tasks and may inspire future research in state space models. However, there could be negative effects brought by the proposed method. For example, the inevitable deviations of training data distribution, the generated results for the real world scenarios may exist the color deviation.

\section{More Results.}
\label{sup_more}

\begin{figure}[h]
  \centering
  \includegraphics[width=0.97\textwidth]{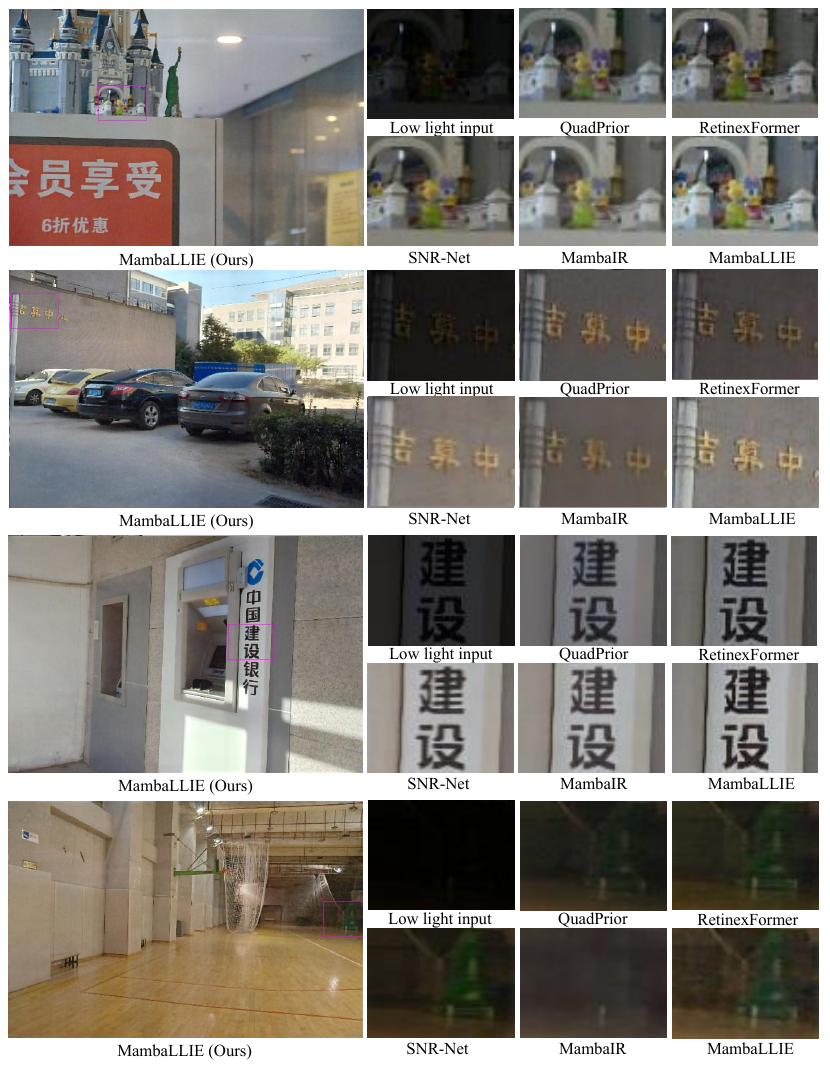}
  \caption{More qualitative comparisons with SOTAs.(Zoom in for best view)}
  \label{fig:append1}
  \vspace{-0.2in}
\end{figure}

\begin{figure}[h]
  \centering
  \includegraphics[width=0.7\textwidth]{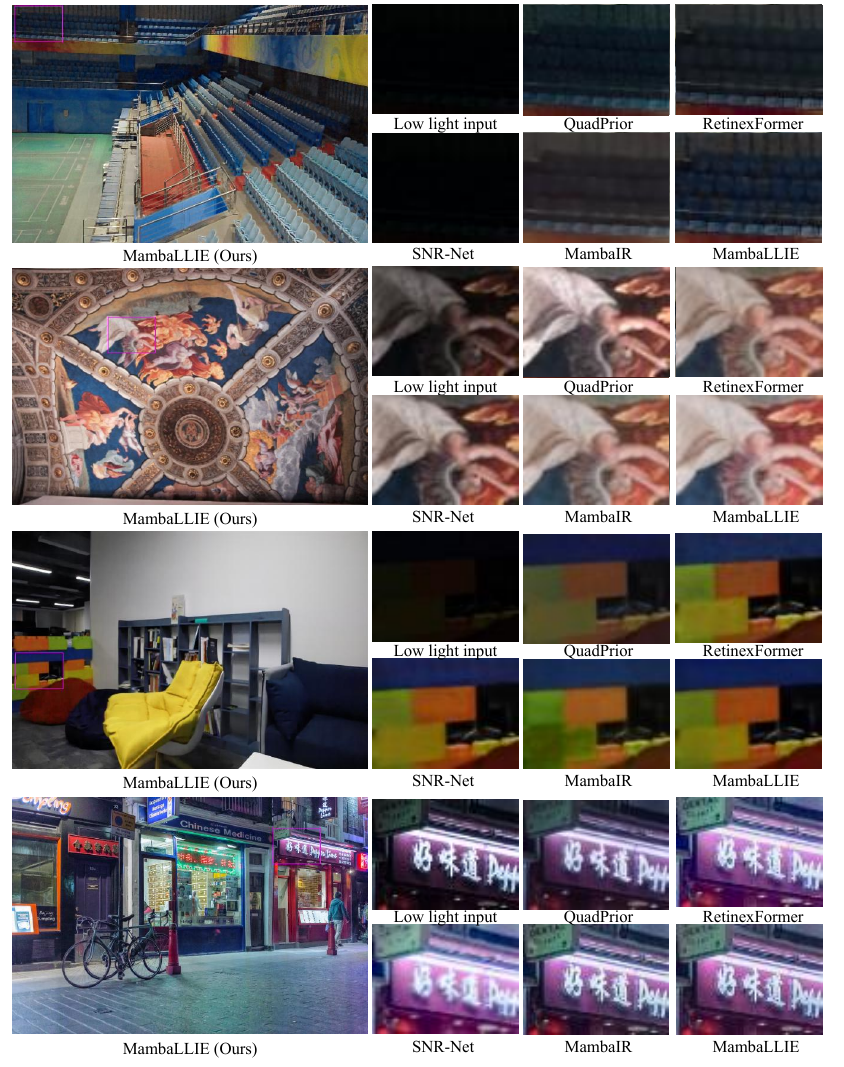}
  \caption{More qualitative comparisons with SOTAs.(Zoom in for best view)}
  \label{fig:append2}
  \vspace{-0.2in}
\end{figure}

\begin{figure}[h]
  \centering
  \includegraphics[width=0.7\textwidth]{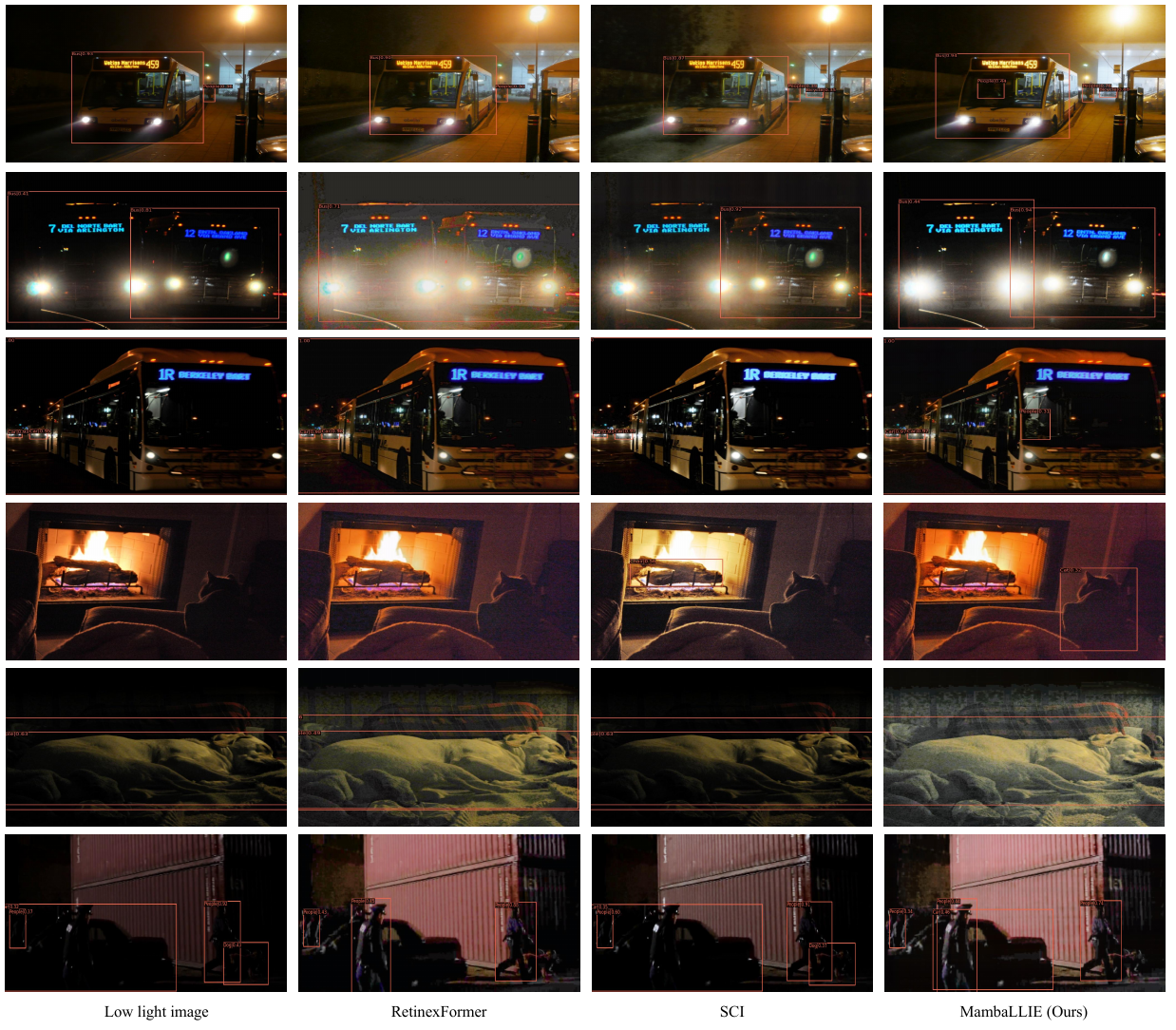}
  \caption{Object detection qualitative comparisons with SOTAs.(Zoom in for best view)}
  \label{fig:appen3}
  \vspace{-0.2in}
\end{figure}

\end{document}